# Multi-tiling Neural Radiance Field (NeRF) – Geometric Assessment on Large-scale Aerial Datasets

Ningli Xu, Rongjun Qin, Debao Huang, Fabio Remondino

*Abstract*— Neural Radiance Fields (NeRF) offer the potential to benefit 3D reconstruction tasks, including aerial photogrammetry. However, the scalability and accuracy of the inferred geometry are not well-documented for large-scale aerial assets,since such datasets usually result in very high memory consumption and slow convergence.. In this paper, we aim to scale the NeRF on large-scael aerial datasets and provide a thorough geometry assessment of NeRF. Specifically, we introduce a location-specific sampling technique as well as a multi-camera tiling (MCT) strategy to reduce memory consumption during image loading for RAM, representation training for GPU memory, and increase the convergence rate within tiles. MCT decomposes a large-frame image into multiple tiled images with different camera models, allowing these small-frame images to be fed into the training process as needed for specific locations without a loss of accuracy. We implement our method on a representative approach, Mip-NeRF, and compare its geometry performance with threephotgrammetric MVS pipelines on two typical aerial datasets against LiDAR reference data. Both qualitative and quantitative results suggest that the proposed NeRF approach produces better completeness and object details than traditional approaches, although as of now, it still falls short in terms of accuracy. The codes are available at https://github.com/GDAOSU/multicamera_nerf.

## I. INTRODUCTION

Photogrammetry is a widely accepted method for 3D scene modeling from convergent images. As a traditional approach [1]–[4], its accuracy and scalability have been well studied in the literature. Typically, photogrammetric reconstruction consists of two components: 1) camera pose estimation or so-called sparse reconstruction through feature-based triangulation, and optionally geo-referencing; 2) multi-view stereo (MVS) based dense image matching for dense point cloud and surface generation. Traditional photogrammetric processing pipelines utilize well-established algorithms and their variants, such as using scale-invariant feature transform (SIFT) or its variants as feature extractor/matcher [5]–[7], as well as semi-global matching (SGM) or patch-based matching [4], [8] as the dense matcher. Although being deployed in various applications, traditional MVS methods suffer from known issues, such as large errors in texture-less regions, reflecting surfaces, and topologically complex structures [9]. Therefore, learning-based methods [10]–[12] have recently been sought, as these methods are data-driven and sufficiently sophisticated to handle scenes where traditional approaches fail. Existing efforts mostly replace certain parts of the traditional MVS pipeline with learning-based components, such as feature extraction [13], depth fusion [14], or multi-view image depth inference [15]. These approaches solve the problems to a certain extent, while such a supervised approach with a complex black-box model not only requires large computation, but also suffers from generalization issues when applied to datasets not seen by the training data [11]. Because of these limitations, unsupervised methods that learn scene representations [16]–[18] from images are considered reasonable alternatives.

As an unsupervised learning approach, Neural radiance field (NeRF) [16], [19]–[21] has recently received increasing attention, due to its ability to learn powerful 3D implicit scene representations used for realistic view synthesis. A typical NeRF model is implemented as a straightforward Multi-Layer Perceptron (MLP), which takes a 5D vector (comprising 3D position and viewing angle) as input and generates a corresponding 4D vector (representing color and density). A NeRF algorithm employs volumetric rendering, which involves accumulating the estimated color and density of samples along rays to synthesize arbitrary views. During optimization, NeRF is trained on a collection of oriented images by minimizing the photometric loss between the rendered view and the input view. This learned MLP model as an implicit 3D representation, provides a means to query color and its density given a 3D location and view direction. Driven by this 3D implicit representation model, existing studies showed that NeRF and its variants can generate consistent views for non-cooperate objects, such as texture-less, transparent, and reflecting surfaces [16]–[18], As a by-product, explicit 3D geometry can be derived from NeRF, which raises a research question: can NeRF, as it does for view rendering, benefit 3D reconstruction from image datasets for photogrammetric purposes?

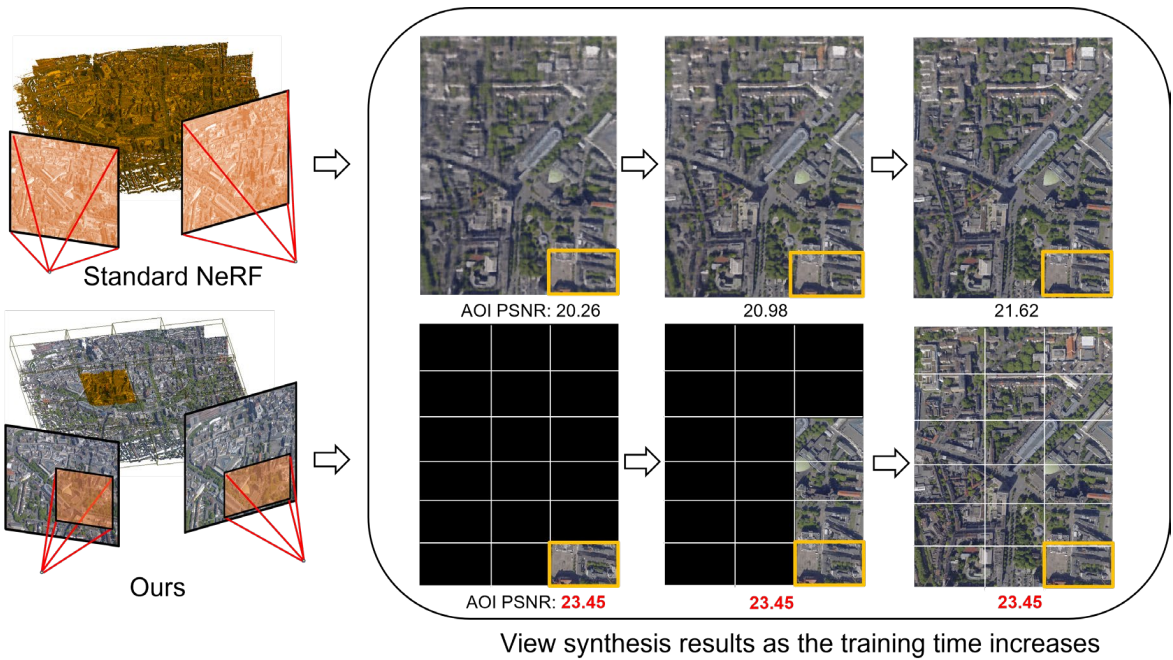

**Fig. 1** Illustration of our location-specific sampling technique (second row) in comparison to the standard sampling strategy (first row). Each column of the right figure denotes the intermediate result after training for a certain time, being 2, 8, and 26 hours respectively using standard NeRF and our method. The standard approach sample pixels across the entire image frame, while ours prioritize areas of interest (AOI) by constraining the sampling within the tiles per epoch, e.g., the rectangle region of the image frame. The AOI PSNR shows that our approach achieves significantly better PSNR in AOIs at the early stage of the training process, while it takes a long time for the standard approach to converge.

This paper aims to answer this question by focusing on enabling NeRF on aerial datasets in a memory-efficient way and assessing the performance of NeRF-derived 3D products. We present a NeRF approach for large-format aerial images that dramatically reduces the needed computing resources for training large scene models. Specifically, we propose a memory-efficient variant of NeRF [16], [17] that (i) effectively reduces the random-access memory (RAM) demands of both data and models, (ii) enables effective training on large-format aerial datasets with affordable graphical processing units (GPUs), and (iii) achieves geometrically detailed and accurate 3D results. Typically, scaling NeRF to large assets requires an out-of-core sampling strategy, which is performed by sampling pixels/rays randomly through the entire image (see top row of Figure 1), which is memory-efficient while ineffective to benefit fast training for specific locations [22], [23]. In contrast, our proposed approach (see second row of Figure 1) involves (i) a location-specific sampling technique that efficiently samples within a smaller sub-region during a certain training duration, and (ii) a multi-camera tiling process that partitions high-resolution aerial images into smaller patches associated to multiple camera models with different intrinsic matrices. These smaller patches are already in the same format as those of common NeRF variants [16], [17], [24], which only requires changing the data loading process to support multiple camera models.

To evaluate our proposed NeRF-based approach, we compute the NeRF-derived 3D geometry and comprehensively compare it with traditional MVS pipelines, including commercial off-the-shelf and open-source software: Agisoft Metashape (https://www.agisoft.com/), OpenMVS (https://github.com/cdcseacave/openMVS, patch-based matching), and a self-implemented MVS algorithm based on SGM (https://u.osu.edu/qin.324/msp/). The evaluation is conducted using two aerial datasets comprising high-resolution aerial images with megapixel resolution, encompassing large scenes as detailed in Table 1. The accuracy of the reconstructed 3D models is assessed using both terrestrial laser scan (TLS) and airborne laser scanner (ALS) data. The comparison includes quantitative and qualitative evaluations across various surface types, such as tiny structures, shadow areas, and texture-less areas. Furthermore, the study analyzes the advantages and limitations of NeRF-based methods for 3D geometry reconstruction.

The remainder of this paper is organized as follows: Section 2 describes the related work of large-scale image-based 3D reconstruction including the photogrammetry-based methods and NeRF-based methods. Section 3 entails the NeRF background and the proposed variant. Sections 4 and 5 describe the experiment settings and results on two aerial datasets, where their performance on the whole and specific regions are detailed. Section 6 summarizes the previous analysis and provides the conclusion.

II. RELATED WORKS

Photogrammetric 3D Reconstruction: Image-based 3D photogrammetric modeling from a set of images has been explored for decades [2], [9], [15], [25]–[30]. Despite that, it remains an open topic that continues to be investigated by computer vision and photogrammetry communities. Nowadays, there exists a series of mature software packages that implement well-engineered algorithms that present the state of the practice. This is particularly true for large-format manned/unmanned aerial and satellite images, as these well-rounded software packages are specifically designed to process image data with large size and volume. Typically, used software packages in the research community, include Pix4D (https://www.pix4d.com/), Agisoft Metashape (https://www.agisoft.com/), SURE (https://www.nframes.com/products/sure-aerial/), and COLMAP (https://github.com/colmap/colmap). Most of these works involve two key components: 1) estimating the camera poses using bundle adjustment [31], [32] and 2) recovering a 3D surface model using MVS algorithms [3], [4], [33]. Most of the MVS methods are based on global, or semi-global optimizations of photometric consistencies across multiple images, such as normalized cross-correlation [34], semi-global matching [4], graph-cut [35], patch-based matching [8], mesh-based geometry refinement [36] and space-carving [37]. These photo-consistency-based approaches, although robust and efficient to implement, suffer from a few known issues: 1) They generate large matching ambiguities at texture-less regions, where photometric features are not sufficiently distinct to yield correct matches. 2) They introduce large outliers at non-cooperative surfaces, such as transparent, or reflecting surfaces, as these photometrics are inconsistent for such surfaces [9]. 3) They suffer from temporally transient images, where the environmental lighting and dynamic objects play as disturbing factors for MVS algorithms. Recent efforts tend to address or alleviate these problems via deep learning-based approaches. The underlying rationale is that, by exploring more sophisticated (and oftentimes black box) models, the complex pixel/feature association can be taught by examples through learning. Typically, these MVS deep learning approaches involve more sophisticated and learned feature metrics (e.g., through the Siamese network [38], [39]), fusion networks that embedded scene priors (planarity and contextual information [40], [41]), etc. Such approaches were shown to be effective to a certain extent, while like many other deep learning methods, suffer from generalization issues, which may perform even worse when applying to datasets that are not seen in training samples [42]. Therefore, unsupervised learning methods that are able to learn sophisticated 3D scene representation without suffering from generalization problems are more favored [43].

Neural Radiance Field: As an unsupervised learning approach, NeRF and its variants [16], [17], [20], [21] are extremely effective in learning implicit 3D representations used for view synthesis. With a set of orientated images as input, it learns the radiance field of the scene using an MLP model, such that the reproduced/rendered views and images can be consistent with the input. Therefore, the MLP represents a field that provides visible color and density given a queried 3D position and viewing direction. Depending on the size of the scene and the resolution of the images, the MLP is built with sufficient parameters to encode details of the scene geometry. NeRF was shown to work well on bounded scenes with well-designed dome collections since bounded scenes with uniformly sampled rays can be easily learned by the MLP. However, when it comes to unbounded scenes, such as for aerial or satellite datasets for outdoor scenes, NeRF may suffer from under-parameterization given the large scene contents and high-resolution images, thus it requires a much larger MLP and video RAM (VRAM, also known as GPU memory) for both training and inferencing. Typical solutions explore hierarchical representations or reduce the scene space by using view-dependent visual frustums. For example, Mip-NeRF [17] represents the scene at continuous scales and render scenes locally to save computing resource. Mip-NeRF360 [18] uses multiple small-sized and granulated MLPs to save model size. Other solutions, such as Mega-NeRF [23] and Block-NeRF [22], spatially tiled the scene and performed independent optimizations. Moreover, the training process of NeRF can be time-consuming, for example, optimizing a standard NeRF on drone datasets takes around 3 days with a cluster of high-performing GPUs [23]. InstantNGP [20] was developed to speed up this process: it uses small-size MLPs with a hash position encoding of 3D points. Among many of these variants, Nerfacto [24] is a relatively more advanced one that combines both the hash encoding from InstantNGP [20], granulated MLP, and scene partitioning from Mip-NeRF360 [18], which are well implemented in open-source packages and achieved state-of-the-art (SOTA) results.

NeRF is known to be particularly successful in view rendering of traditionally difficult and non-cooperative objects, such as texture-less, transparent, and reflecting surfaces [57]. However, it is still unclear how the NeRF-derived 3D geometry performs on large-sized aerial images, partly due to the scale of the problem. An earlier work [44] evaluated the performances of various NeRF methods on close-range heritage assets, showing that NeRF-derived 3D geometry can be robust to reflective and transparent surfaces. Therefore, we expect that a thorough evaluation of aerial scenarios can be particularly useful to assess the NeRF-derived 3D geometry for mapping.

## III. METHODOLOGY

We aim to compute NeRF-derived 3D geometry on full-scale aerial datasets and evaluate its geometric reconstruction accuracy against traditional photogrammetric methods. Since NeRF requires oriented images, we first perform a traditional bundle adjustment on the aerial images and supply the same poses to NeRF and typical MVS pipelines for computations. Section 3.A entails the basic concepts and the approach used to derive 3D geometry. Then, Section 3.B introduces the proposed NeRF variant for large-scale aerial datasets.

### A. Neural radiance filed (NeRF)

NeRF is a coordinate-based neural scene representation that learns to model the scene appearance by optimizing the photometric loss of reproducing the appearance of a set of oriented images. A fully optimized NeRF can then be used to render from arbitrary views, including previously unseen viewpoints of the scene.

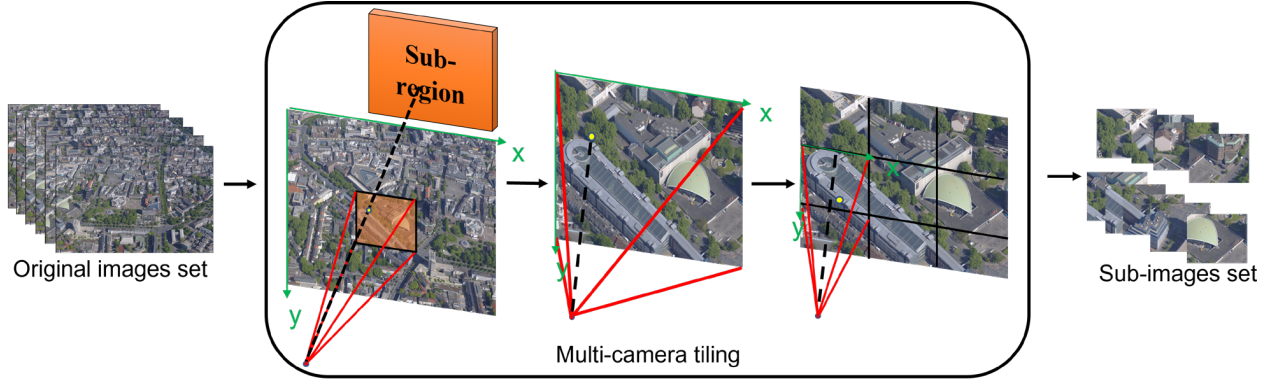

**Fig. 2** The proposed multi-camera tiling process, which extracts a set of sub-images corresponding to the targeted sub-region. The detailed explanation can be found in Section 3.B.2.

Standard NeRF [16] applies a pair of MLPs to model the scene appearance which takes a 5D vector (comprising 3D position and viewing angle) as input and generates a corresponding 4D vector (representing color and density). The first MLP $f_\sigma$ takes in a 3D position $x$ and outputs volume density $\sigma$ along with a feature vector. This feature vector is concatenated with a 2D viewing direction $d$ as the input of the second MLP $f_c$ that outputs the color $c$. The underlying geometry of the scene is modeled in volume density which is the function of 3D position. The modeling of color incorporates viewing direction which allows NeRF to represent non-cooperate objects, such as reflective or transparent surfaces.

The color of each pixel can then be calculated by accumulating the color and density of samples along the corresponding ray in 3D space. Specifically, each pixel corresponds to a ray $r(t) = o + td$ in the 3D space. NeRF $f_\theta$ uniformly samples distances $\{t_i\}_{i=0}^N$ along the ray and passes the 3D points $r(t_i)$ and direction $d$ through its MLPs to calculate the $\sigma_i$ and $c_i$. The color of each pixel $c_{pix}$ is calculated as Eq1:

$$c_{pix} = f_\theta(r) = \sum_{i=1}^{N} w_i c_i, \text{where } w_i = T_i(1 - e^{-\triangle_i \sigma_i}) \quad (1)$$

$$T_i = \exp\left(-\sum_{j<i} \triangle_j \sigma_j\right), \quad \triangle_i = t_i - t_{i-1} \quad (2)$$

The value of $c_{pix}$ is a weighted average of sampled color density at the ray direction, and it penalizes occluded points based on the accumulated color density (Eq2). By optimizing the total squared error of the rendering and the true pixel colors (Eq3), NeRF can be trained through stochastic gradient descent algorithms. At each iteration, NeRF performs the back-propagation based on a batch of camera rays $r$ randomly sampled from a set of pixels $\mathcal{R}$.

$$\mathcal{L} = \sum_{r \in \mathcal{R}} \|f_\theta(r) - c(r)\|_2^2 \quad (3)$$

The pixel color is ideally an integration of all incoming radiances within the pixel's frustum while standard NeRF only integrates the samples along the ray centered at the pixel. Mip-NeRF [17] extends the way of standard NeRF calculating pixel color by integrating the samples within the conical frustums rather than rays. In practice, it approximates these frustums through Gaussian modeling [17].

**Memory Consumption:** Memory usage becomes a critical factor as the dataset size (image size and number) expands. The primary role of RAM is to temporarily store full-size images for processing (termed as #cached images in Table 1). In practice, it is possible to perform out-of-core training, by selecting the limited number of rays for training at a time. VRAM (or GPU memory), on the other hand, is responsible for providing the necessary space for inferences and back-propagation of gradients concerning the learnable parameters. Throughout the training process, the extent of memory consumption is closely tied to the size of the training batch, which encompasses both the sampled camera rays $r$ (termed as #camera rays in Table 2) and the number of samples along each ray $\{t_i\}_{i=0}^N$ (termed as #samples in Table 2).

**Deriving 3D Geometry:** The depth of a single ray can be calculated by the weighted sum of the location of the sample as shown in Eq-4, where the weight $w_i$ is the product of the accumulated transmittance and the local density as described in Eq-1 denoting the chance of the ray terminating at the current location [45]. Then, the dense point cloud of the whole scene is the combination of the depth maps of all oriented images. Moreover, the 3D surface mesh model can be derived by first combining the depth map of each view to construct the truncated signed distance function (TSDF) [46] over 3D voxel grid, and then using the marching cubes method [47] to examine each voxel grid to approximate the surface of the object.

$$d = \frac{\sum_{i=0}^{N} w_i t_{i-0.5}}{\sum_{i=0}^{N} w_i}, \quad t_{i-0.5} = \frac{(t_i + t_{i-1})}{2} \quad (4)$$

*B. Enabling NeRF on large-scale aerial datasets*

Incorporating NeRF into large-scale aerial datasets remains challenging due to issues such as slow convergence and high memory consumption. Training NeRF on large aerial datasets typically demands a considerable amount of time, often spanning around a week [23]. Additionally, effectively modeling a large-scale scene necessitates an MLP with millions of learnable parameters, leading to substantial GPU memory requirements for parameter updates. Moreover, handling megapixels of aerial imagery proves to be a daunting task, as it becomes challenging to process or even load into RAM.

We address these issues by proposing a novel approach that efficiently stores the pixel-wise information according to the spatial location. Specifically, our approach introduces a mechanism for targeted sampling based on location, called **location-specific sampling** (detailed in Section 3.B.1). This mechanism empowers NeRF models to systematically iterate through sub-regions of the whole scene and allocate all the training resources when training with each sub-region. To facilitate the integration of this sampling mechanism into existing NeRF methodologies, we propose a **multi-camera tiling (MCT)** technique (elaborated in Section 3.B.2). This technique involves subprocesses such as location-specific view indexing, image reshaping and intrinsic parameters updating.

*B.1. Location-specific sampling*

We observe that as we scale up the size of datasets, it becomes evident that the commonly used random sampling mechanism in NeRF loses its efficiency in progressive visualization. At each training iteration, standard NeRF randomly samples from the whole pixel set $\mathcal{R}$ (in Eq-3) to optimize the radiance field. Its large sampling range makes the sampled pixel density (number of sampled pixels per object space unit) dramatically small, leading to blurry view synthesis results during the training process. To address this issue, we introduce a **location-specific sampling** mechanism, which constrains the sampling range within a smaller object space $O_{small}$. Such constraints in object space can be easily projected to pixel space, according to Eq-5, where $[X, Y, Z, 1]^T \in R^{4 \times 1}$ represents object location at homogeneous coordinates; $[x_{pix}, y_{pix}, 1]^T \in R^{3 \times 1}$ is the corresponding pixel location; $P \in R^{3 \times 4}$ is the projection matrix containing both intrinsic and extrinsic camera parameters. By projecting the boundary coordinates of $O_{sub}$, we can calculate their pixel coordinates and thus the pixel set $\mathcal{R}_{small}$.

$$\begin{bmatrix} x_{pix} \\ y_{pix} \\ 1 \end{bmatrix} = P \begin{bmatrix} X \\ Y \\ Z \\ 1 \end{bmatrix} \quad (5)$$

Given the bounding box of the entire scene, our approach first partitions the scene into several sub-regions, iterates through each sub-region, and uses location-specific sampling within the sub-region. To achieve this, we follow the concept of the Neural Ground Plan (NGP) [48] that assumes the scene can be represented as a flat surface and partition it into a two-dimensional grid from a top-down perspective. Instead of optimizing the whole scene, ours only optimizes a certain sub-region at each time with the location-specific sampling technique, which largely reduces the sampling range and thus improves the training efficiency. In practice, we expand the boundary of each sub-region with a marginal overlap (e.g. a factor of 2 for lateral sub-region size is used) to avoid occlusion and visibility issues.

*B.2. Multi-camera tiling*

To support location-specific sampling in existing NeRF methods, we build data structures that allow adaptive indexing of relevant images of a given location. Existing methods, such as Mega-NeRF [23], perform ray casting to calculate the distance of every camera ray to the sub-region centroid and allocate pixels that are close enough to that sub-region centroid, which requires a high memory capacity (statistics shown in Table 3). Instead of using ray casting, we project the sub-region location in object space back to pixel space to determine the qualified sub-image coordinates (as depicted in Eq-5). In practice, we found using the sub-image format with multiple camera models is efficient in representing such pixel sets.

Specifically, we propose an efficient tiling technique, called **multi-camera tiling (MCT)**, which is applied to the original images set $I$ to generate the sub-images set $I_i$ (as shown in Figure 2). It crops out the sub-image based on the corresponding sub-image region using Eq-5. The principal points $(C_x, C_y)$ are updated for each tile. The cropped image still shares the same focal length and center perspectives, as well as the same rotation matrix. The update of principal points is straightforward and can be found in Eq-6:

$$\begin{aligned} C_{x_{new}} &= C_x - O_{x_{new}} \\ C_{y_{new}} &= C_y - O_{y_{new}} \end{aligned} \quad (6)$$

where $(C_x, C_y)$ is the pixel location of the principal point for the original image, and $(C_{x_{new}}, C_{y_{new}})$ is the pixel location of the principal point for the tiled images. $(O_{x_{new}}, O_{y_{new}})$ is the pixel location of the origin of the new coordinates in the original image coordinate system. With the tiled images, the minimum RAM requirement can be significantly reduced, particularly for high-resolution images. For instance, using an image of 8176 x 6132 pixels, the theoretical RAM consumption would be approximately 143MB. When the size of the sub-scene is so small that the sub-image size is around 800 x 800, the memory consumption is brought down to 1.83MB.

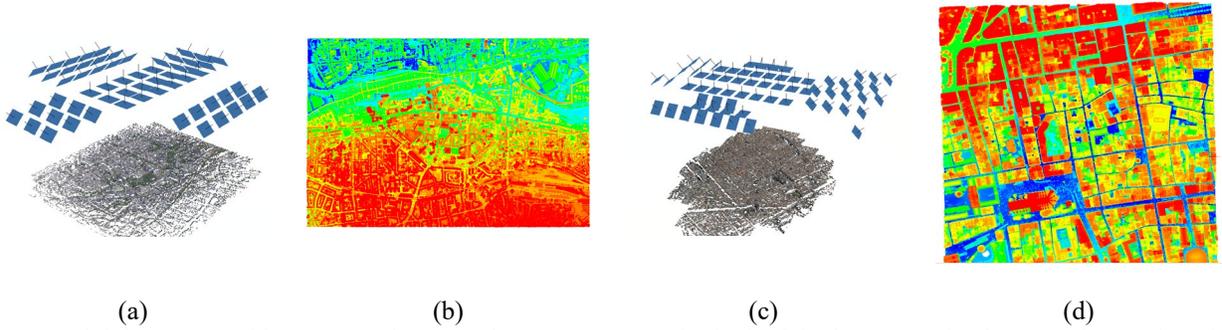

(a) (b) (c) (d)

**Fig. 3** Two aerial datasets used in our experiments. The camera networks (a,c) of the images and reference LiDAR data (b,d) for the Dortmund and Bordeaux datasets.

**Table 1** Detailed information about two aerial datasets used in our work. "TLS" represents terrestrial laser scanners. "ALS" represents the airborne LiDAR data.

| Dataset | AOI | Imagery | | | LiDAR | |
|---------|-----|---------|---|---|-------|---|
| | | #Images (nadir/oblique) | Resolutions | GSD | Type | Mean point density |
| Dortmund | 1.5 x 1.5 km$^2$ | 16/43 | 6132 x 8176 pixels (nadir) 8176 x 6132 pixels (oblique) | 10-14 cm | TLS, ALS | 10pts/m$^2$ (TLS) |
| Bordeaux | 1.5 x 1.2 km$^2$ | 33/37 | 10336 x 7788 pixels (Nadir and oblique) | 5 cm | ALS | 22pts/ m$^2$ |

To further enable NeRF to model large outdoor scenes, we follow a representation tiling ideas typically used for approaches with improved scalability, such as from Mega-NeRF [23] and Block-NeRF [22], which, model each sub-scene by an individual radiance field. This spatial partition allows for a set of simple and smaller NeRF models to efficiently represent a large-scale scene. Each NeRF model can be trained using less memory to represent the scene, cast rays, and optimize the parameters. In concept, these techniques can be adapted to an NeRF variants, here in this paper we select Mip-NeRF [17]. The selection of Mip-NeRF is based on the preliminary experiment on aerial datasets including InstantNGP [20] and Nerfacto [24], while the InstantNGP and Nerfacto completely failed. The architecture of Mip-NeRF is similar to standard NeRF [16], with a base MLP that consists of 8 hidden layers with 256 neurons each. In addition, Rather than sampling points along the rays, the rendering of Mip-NeRF is based on integrating volume within conical frustums, which significantly reduces objectionable aliasing artifacts.

## IV. DATASETS

Two aerial photogrammetry datasets (Table 1 and Figure 3) are considered to evaluate the proposed NeRF method in comparison to the baseline methods including Mip-NeRF[17], Mega-NeRF[23] and three traditional MVS pipelines (detailed in Section 5).

**Dortmund dataset** [49]. The ISPRS/EuroSDR benchmark provides multi-camera aerial images captured over the city of Dortmund, Germany. The ground sample distance (GSD) of the nadir and oblique images ranges from 10 to 14cm, respectively. The dataset includes also airborne and ground LiDAR data as a reference for accuracy assessment. For our experiments, a sub-block of 59 images covering an area of approximately 2x2 km$^2$ was chosen (Figure 3-a,b).

**Bordeaux dataset** [50]. This dataset includes concurrent collection of aerial photogrammetry and LiDAR data over the city of Bordeaux, France, using a Leica CityMapper hybrid sensor. The photogrammetric data contains 480 images with a mean GSD of ca. 5 cm whereas the LiDAR data have a mean single strip point density of around 10 pts/m$^2$. For our tests, a sub-block of 70 images covering an area of approximately 1.5 x 1.2 km$^2$ was chosen (Figure 3-c,d).

## V. EXPERIMENTS

We evaluated the accuracy of the generated geometry at the point cloud level. For both datasets, the aerial triangulation of the images was done using Agisoft Metashape, and the images were undistorted using the obtained lens distortion parameters. All experiments were run with full image resolution. To evaluate the quality of the resulting point clouds, we employed the cloud-to-cloud distance measurement [51], using the available LiDAR data as the reference. Additionally, we evaluated the completeness and accuracy of the point clouds by determining the percentage of points falling within varying threshold values [35-36]. All experiments were conducted utilizing an Intel(R) Xeon(R) W-2235 CPU @ 3.8GHz processor, 64GB of RAM, and an NVIDIA GeForce RTX 3090 with 24GB of VRAM.

**Table 2** Essential hyper-parameters of SOTA NeRF methods related to RAM and VRAM. "#camera rays", and "#samples" are detailed in Section 3.A. "\" for Mip-NeRF represents it optimizes over the whole scene. Note: "ours" method is our method based on Mip-NeRF and the "Mip-NeRF" is the original approach.

| Hyper-parameters | Mip-NeRF | Mega-NeRF | Ours |
|---|---|---|---|
| Down-sample factor | 1 | 2 | 1 |
| #sub-regions | \ | 16 | 16 |
| Cached image data | Full frame images | Camera rays/pixels | Image patches |
| #camera rays | 5000 | 5000 | 5000 |
| #samples (rough, fine) | 64,128 | 64,128 | 64,128 |

**Table 3** RAM and VRAM consumption of SOTA NeRF methods. "VRAM" measures the total usage of video RAM (or GPU memory) and the "RAM" is the current operation system RAM usage. "\" for Mip-NeRF means that it directly performs training without any data partition process.

| Method Module | Memory Type | Mip-NeRF | Mega-NeRF | Ours |
|---|---|---|---|---|
| Data Partition | VRAM | \ | 17.7 GB | 0.0 GB |
|  | RAM | \ | 24.5 GB | 1.2 GB |
| Training | VRAM | 8.5 GB | 9.9 GB | 7.2 GB |
|  | RAM | 6.0 GB | 1.8 GB | 2.5 GB |

**Table 4** The convergence rate comparison.

| Training Time [hours] | PSNR | | |
|---|---|---|---|
|  | Mip-NeRF | Mega-NeRF | Ours |
| 6 | 18.36 | 17.06 | 19.06 |
| 12 | 18.90 | 17.76 | 19.86 |
| 36 | 19.56 | 17.95 | 20.94 |

We first evaluated the memory consumption and convergence performance of our approach compared to two SOTA NeRF methods: Mip-NeRF [17] and Mega-NeRF [23]. Then, the best-performed NeRF method was compared in terms of geometric reconstruction performance with three traditional MVS software:

**Agisoft Metashape** (https://www.agisoft.com/) is a professional photogrammetry software used for generating 3D models from a set of 2D images. It employs structure-from-motion (SfM) and MVS to analyze the overlapping images and extract 3D information. Since it is a commercial package, the information on used SfM and MVS algorithms is missing.

**OpenMVS** (https://github.com/cdcseacave/openMVS) is an open-source software package for patch-based MVS. It is based on the concept introduced by [8], which involves initially selecting stereo pairs for each image based on factors such as the viewing angle of visible points and the distance between camera centers. Depth maps are then computed for each pair using a patch-based method [3], followed by a depth refinement process to ensure consistency across neighboring views or priors to improve completeness in texture-less areas [9]. Subsequently, a depth merging process is employed, which considers redundancy and occlusion checks among neighboring images to generate the final photogrammetric point cloud. An extension based on plane priors was presented in [9]

**Multi-view Stereo Processor (**MSP) [25] is a SGM-based MVS algorithm. A set of stereo pairs is first selected for each image based on the criteria including camera poses and number of correspondences. Then, for each image, a Census-based SGM method [4][54] is applied to generate pairwise depth maps for the corresponding stereo pairs, followed by a median filtering method to derive the high-quality per-view depth map. The final photogrammetric point clouds are the merging results of the per-view depth maps.

### A. Evaluations on the entire datasets

#### A.1. State-of-the-Art comparison

The proposed approach was compared to two SOTA NeRF methods (Mip-NeRF [17] and Mega-NeRF [23]) to demonstrate the effectiveness of the proposed NeRF variant for reconstructing large-scale aerial scenarios. We adjusted the essential hyper-parameters related to RAM and VRAM such that they were maximally consistent across all three methods (Table 2). To avoid memory issues caused by Mega-NeRF (crashed in our experiment if original resolutions are used), we down-sampled the image by a factor of two, for this particular comparison (but the full resolution was experimented in full accuracy analysis in Section 5.B).

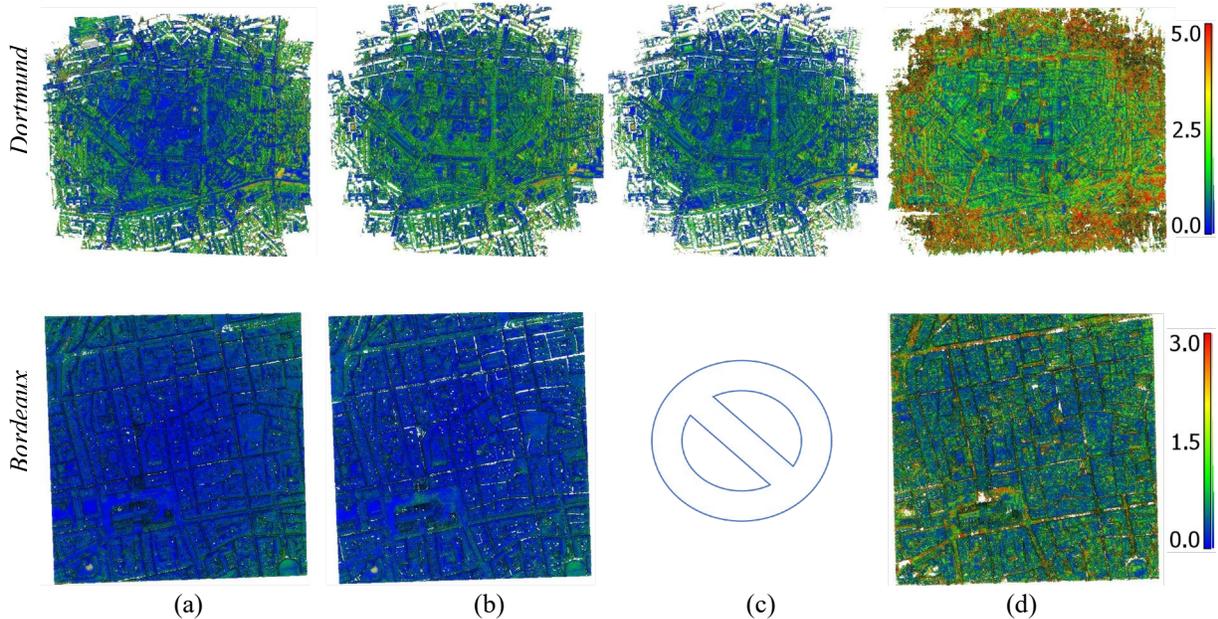

**Fig. 4** Color-coded cloud-to-cloud comparisons for dense point cloud from each method to LiDAR data [unit: m]. (a-d) are results of Metashape, OpenMVS, MSP, and our NeRF method on the Dortmund (top) and Bordeaux (bottom) datasets. For the Bordeaux dataset, all dense point clouds are cropped according to the available LiDAR data boundary.

**Table 5** Metrics of cloud-to-cloud comparison for our NeRF method and three photogrammetry methods [unit: m]. The "mean error" refers to the average absolute distance of the nearest neighboring points. As explained in the text, OpenMVS failed at the Bordeaux dataset.

|  | Dortmund | | Bordeaux | |
| --- | --- | --- | --- | --- |
|  | **Mean Error** | **STD** | **Mean Error** | **STD** |
| **Metashape** | 1.1642 | 1.2296 | 0.3826 | 0.3930 |
| **OpenMVS** | 0.8620 | 1.0808 | Failed | Failed |
| **MSP** | 0.7610 | 1.1563 | 0.2156 | 0.2344 |
| **NeRF (Ours)** | 1.4392 | 1.1916 | 0.9232 | 1.0829 |

The main processes that consume the VRAM were the forward propagation of the training batch and the backward passing of the gradients of learnable parameters. Using the same model architecture and training batch size, our method consumed significantly less VRAM thanks to our MCT strategy. In addition, RAM played an important role in caching the training data. During the training process, As shown in Table 3, the proposed method and Mega-NeRF required much less RAM than Mip-NeRF. This was due to that both ours and Mega-NeRF had a data partition module that allowed an out-of-core process. Moreover, in the data partition process, the memory consumption of Mega-NeRF was predominantly associated with its ray casting procedure, which involved the creation of 3D volumes for each full-size image, while our method performed these steps locally, which can be demonstrated in the "Data Partition" result of Table 3. However, a substantial distinction arised when considering the loading of images into RAM. Unlike Mip-NeRF, which necessitates loading a set of full-size images, our method solely loads the sub-image set. Table 4 demonstrates that, despite having the same training time, our method achieved a superior Peak Signal-to-Noise Ratio (PSNR) compared to Mip-NeRF and Mega-NeRF. This improvement in PSNR suggests that our method achieved a faster convergence rate, even though we employed the same network architecture as Mip-NeRF.

*A.2. Cloud-to-Cloud comparison*

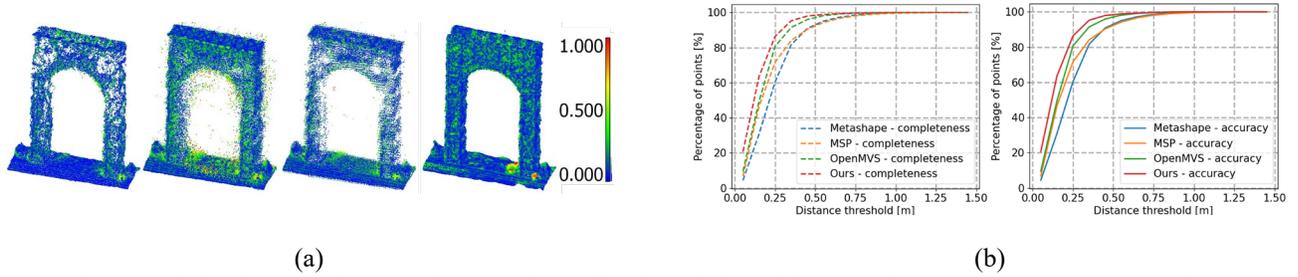

(a)                                                            (b)

**Fig. 5** Color-coded cloud-to-cloud comparison results (a) for, from left to right, Metashape, OpenMVS, MSP, and NeRF method for the fourth object in Figure 6. Completeness and accuracy curves (b).

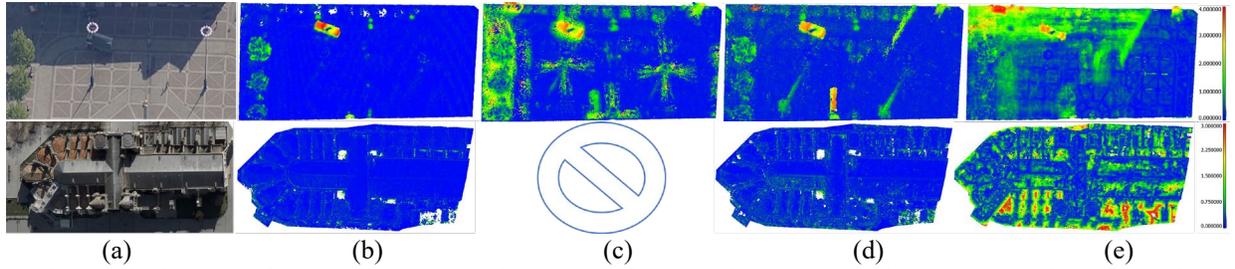

(a)             (b)             (c)             (d)             (e)

**Fig. 6** Selected shadow areas in Dortmund (top) and Bordeaux (bottom) dataset as seen in the camera views (a). Color-coded cloud-to-cloud comparison results for Metashape (b), OpenMVS (c), MSP (d), and NeRF method (e). Note OpenMVS failed at the Bordeaux dataset (explained in Section 5.A.1), thus the result is not included.

**Table 6** Metrics of cloud-to-cloud comparisons for the considered methods [unit: m] (see Figure 8).

|                     | **Building (Bordeaux)** |         | **Ground (Dortmund)** |       |
|---------------------|:-----------------------:|:-------:|:---------------------:|:-----:|
|                     | **Mean Error**          | **STD** | **Mean Error**        | **STD** |
| **Metashape**       | 0.181                   | 0.237   | 0.151                 | 0.325 |
| **OpenMVS**         | Failed                  | Failed  | 0.175                 | 0.344 |
| **MSP**             | 0.126                   | 0.113   | 0.153                 | 0.255 |
| **Our NeRF method** | 0.456                   | 0.541   | 0.475                 | 0.472 |

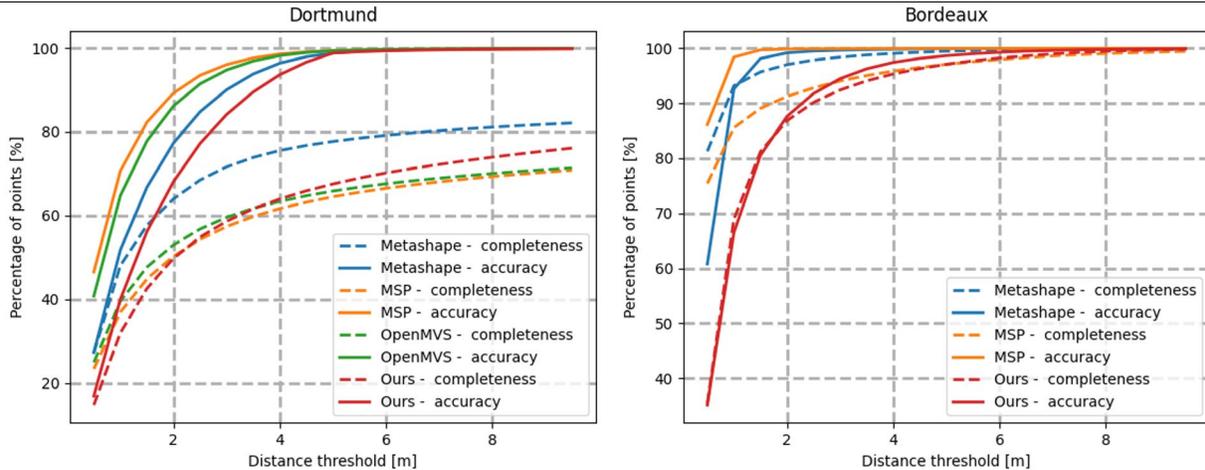

**Fig. 7** Accuracy and completeness of our NeRF method and three photogrammetry methods in Dortmund and Bordeaux datasets. As noted in the text, OpenMVS failed for the Bordeaux dataset and thus is not included.

A cloud-to-cloud comparison refers to the measurement of absolute Euclidean distances between 3D samples in a dataset concerning the reference data [51], [55]. For both aerial datasets, dense point clouds were derived using three photogrammetry methods (OpenMVS, MSP, Metashape) as well as the proposed NeRF variant, then co-registered to the available ground truth LiDAR data (Figure 4) and comparison metrics derived (Table 5). It should be noted that OpenMVS failed to generate point clouds for the Bordeaux dataset due to memory overflow in the depth fusion process. Figure 4 illustrates that our NeRF method yielded less accurate results, where the inaccurate points are mainly located near the boundary in the Dortmund dataset due to the lack of images at the collection boundaries. This is a known issue with NeRF when using sparse views, causing inaccurate points near the camera centers[56], [57]. Table 5 confirms that NeRF had the highest mean errors (1.4392m in Dortmund and 0.9232m in Bordeaux). In comparison, MSP had the lowest mean error (0.7610m), followed by OpenMVS (0.8620m) and Metashape (1.1642m) in the Dortmund dataset.

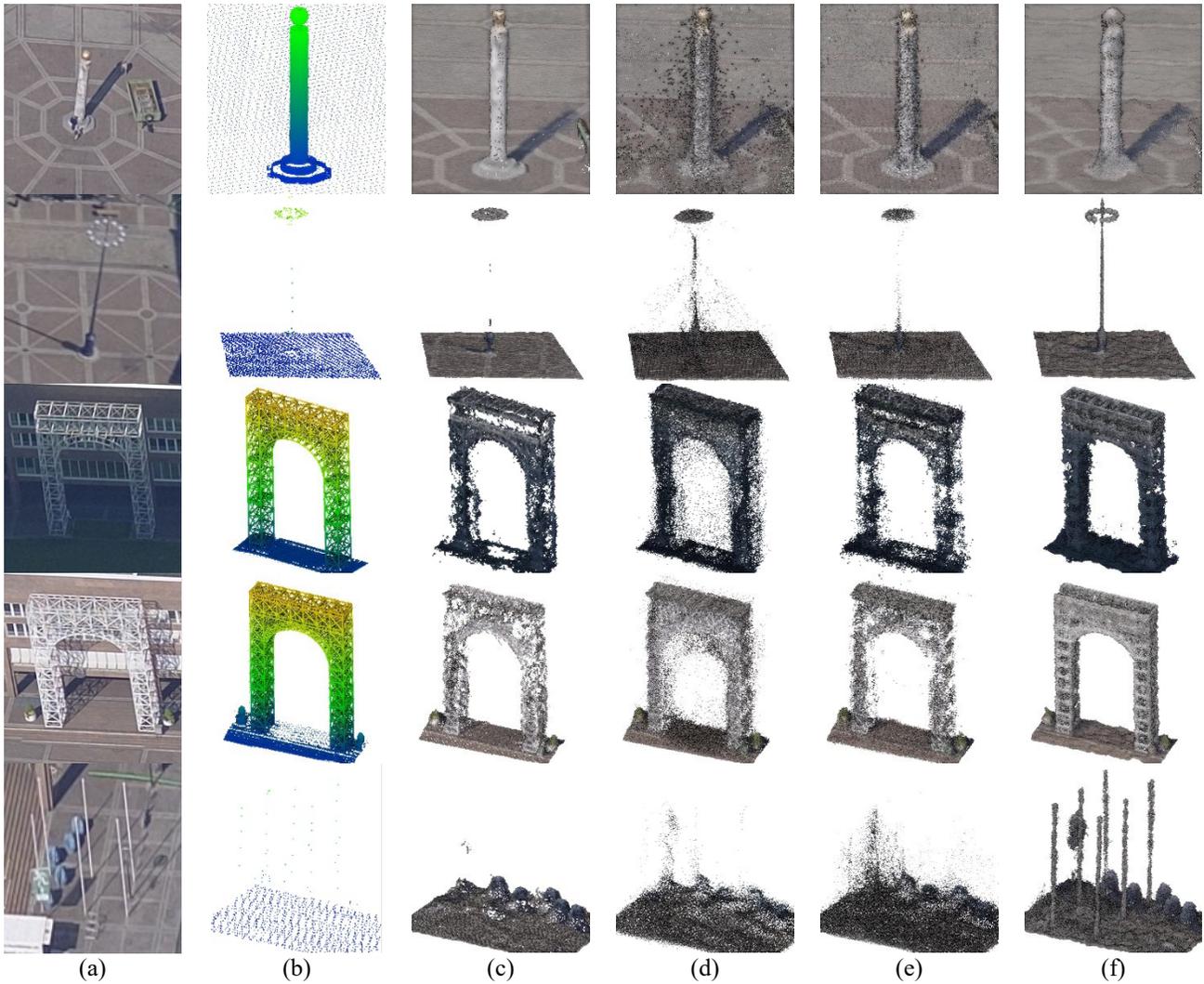

**Fig. 8** Different fine structures in the Dortmund area (a), ground truth LiDAR data (b) and achieved results with the different methods: Metashape (c), OpenMVS (d), MSP (e) and proposed NeRF method (f). The first row is a stone pillar of 7-11 pixels in width; the second row is a light bulb of 4-5 pixels in width; the third and fourth row is the steel bar of 2-3 pixels in width; the fifth row is the stone pillar of 2-3 pixels in width.

*A.3. Accuracy and completeness*

The accuracy and completeness of the aforementioned 3D reconstruction outcomes were evaluated using varying distance thresholds. Accuracy refers to the percentage of examined point clouds deemed accurate (falling within a specific distance threshold from the LiDAR data) while completeness is the percentage of LiDAR points that are covered by the examined point cloud (falling within a specified distance of the examined point cloud). The findings, illustrated in Figure 5, validate that NeRF consistently yielded less precise results than any of the photogrammetry methods, regardless of the chosen distance threshold. Examining the completeness curve for the Dortmund dataset, we observed that NeRF achieved comparable completeness to MSP and OpenMVS when the threshold was set at less than 3 meters whereas NeRF outperforms these two photogrammetry methods in terms of completeness when the threshold exceeded 3 meters. Notably, the points contributing to this improved completeness were primarily those labeled as "yellow" and "red" in Figure 4-d.

*B. Evaluations on the selected regions*

*B.1. Fine structures*

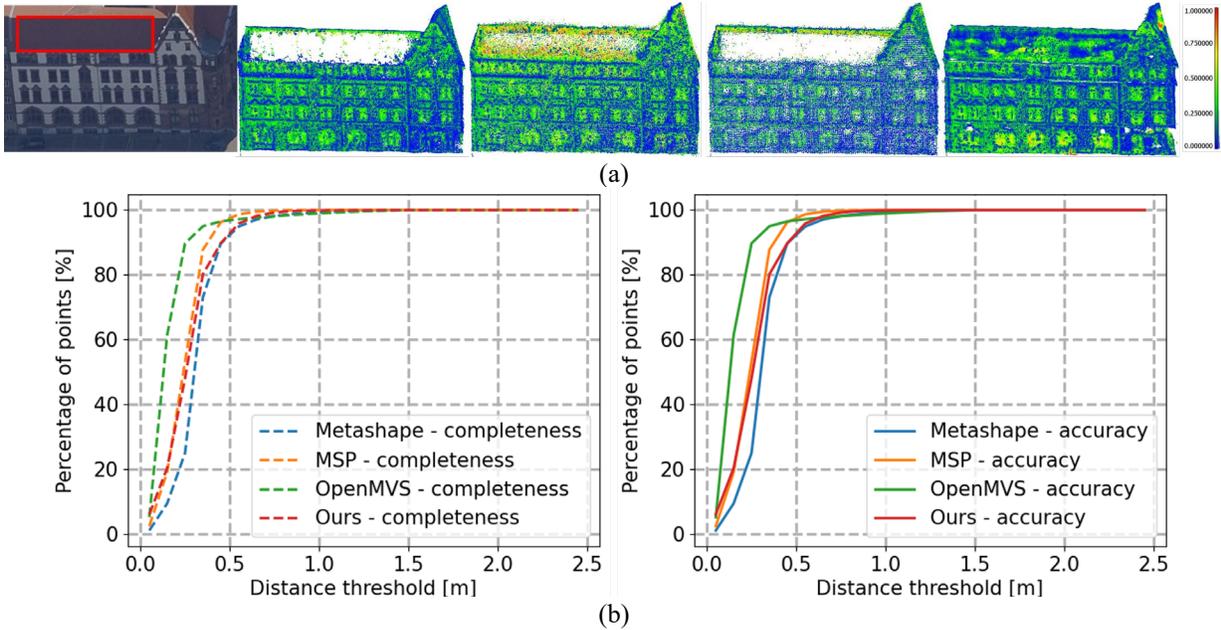

**Fig. 9** The selected texture-less area (Dortmund dataset) and, (a) from left to right, color-coded cloud-to-cloud comparison result of Metashape, OpenMVS, MSP, and the NeRF method. (b) Completeness and accuracy curves of results.

Traditional MVS pipelines often encounter difficulties in reconstructing small and/or thin objects due to the imposed smoothness constraints, which penalize depth discontinuities within local surfaces. This penalty ultimately leads to the loss of fine details and information. Figure 6 presents some tiny objects which vary in width from 2 to 11 pixels in the image space and results show that the proposed NeRF method outperformed traditional MVS pipelines in terms of completeness. For instance, the NeRF method successfully reconstructed the thin light pole in the second row, the steel bars in a horizontal direction in the fourth row, and the six stone pillars in the fifth row, none of which can be reconstructed by traditional MVS pipelines. As the width of objects increases to 7-11 pixels of footprint in images, both photogrammetry and NeRF methods can reconstruct them with completion (Figure 6 first row).

A quantitative analysis of the object in the fourth row of Figure 6 is shown in Figure 7. NeRF exhibited the highest completeness while maintaining comparable accuracy to traditional MVS pipelines. Within a tolerance of 0.25m, the NeRF successfully reconstructed 86.3% of the LiDAR data, followed by OpenMVS at 80.7%, MSP at 71.6%, and Metashape at 60.4%. Notably, the largest discrepancy is observed in the completeness curve and the lower part of the object.

*B.2. Shadow areas*
Real-world aerial datasets often contain shadow areas, which are of particular interest for our analysis. To represent such regions, we selected two distinctive areas: the first comprised a square ground, half of which was covered in shadow while the other half was under direct sunlight. The second area was a church building with its north side in shadow, as depicted in Figure 8. The 3D results obtained by our NeRF method exhibited notable accuracy inconsistencies between the shadow and sunshine areas, whereas traditional methods yielded more consistent outcomes. By looking at the images in Figure 8-a, the shadow area had a similar texture pattern as the sunshine area while having a different level of illustration. Moreover, even in the non-shaded area, our NeRF method performed worse in such flat surfaces, especially in areas with uniform color patterns. To quantitatively assess the accuracy, mean and standard deviation errors with respect to the LiDAR data were computed (Table 6). Combining the visual and quantitative findings, we observed that NeRF produced less accurate geometry on flat surfaces, and their performance further deteriorated when image intensity decreased, primarily due to that the loss was built based on the photometric loss, which was less informative for pixel at low intensity.

*B.3. Texture-less areas*
In traditional MVS pipelines, it is common to encounter low completeness in 3D results, particularly in texture-less regions present on building facades and water surfaces. This issue arises due to the significant matching ambiguity in such areas. We selected a building displaying a uniform color pattern on its roof, captured in more than 30 images. Figure 9-a demonstrates that the NeRF method effectively filled the holes in the textureless surface whereas traditional MVS pipelines, except OpenMVS, failed to do so.

While our NeRF method produced visually complete results (less emptiness), the completeness curve in Figure 9-b indicates that it underperforms OpenMVS. This suggests that the NeRF method generally introduced more erroneous points. The non-hole area primarily represented a flat surface under shadow conditions, where the NeRF method exhibited inferior performance, as explained in the previous section.

*C. Analysis and summary*

In summary, the proposed strategy enables NeRF (hereafter we called our NeRF method for simplicity) to achieve better convergence rate and requires less RAM and VRAM than the original version of SOTA methods for aerial cases. Its derived 3D results underperform traditional MVS pipelines in terms of accuracy, particularly in shadow areas. However, it demonstrated better performance at reconstructing small objects (i.e., complex structures with parts taking small number of pixels in the image) and texture-less regions. Specifically:

**Advantages:**
- Our NeRF method demonstrates improved training convergence rates due to the location-specific sampling strategy. Additionally, it requires significantly less RAM during data partition process, due to the proposed multi-camera tiling technique.
- Our NeRF method excels in reconstructing intricate geometric structures that are challenging for traditional MVS pipelines, such as thin light bulb pillars and steel pillars. Unlike traditional pipelines, NeRF applies a per-pixel photometric loss function that does not penalize depth discontinuity between neighboring pixels.
- On processing the aerial blocks, our NeRF method (generally for NeRF approaches) generates denser and visually more complete geometry. Traditional MVS pipelines produce holes due to depth estimation based on limited observations from neighboring images and subsequent outlier removal processes. In contrast, NeRF estimates geometry by optimizing a single cost function that incorporates all multi-view observations.

**Disadvantages:**
- Geometry produced by our NeRF method often exhibits errors in shadow areas. This problem can be attributed to the instability of backpropagation during the optimization process, due to that the loss was constructed based on the pixel intensity. Moreover, the depth generation process of NeRF (Eq-1,2,4) is a function of the color intensity. Therefore, the depth exhibit certain correlation with the brightness of the pixels, and introduce unwanted errors, which are very often visible for in depth maps of flat regions where brightness of the texture varies.
- The geometry generated by NeRF is generally than that produced by traditional MVS pipelines. Experimental results indicate that the NeRF generated results possess larger geometric uncertainties, often observed at flat regions. NeRF operates its cost on distributions based on photo-consistency, while traditional MVS pipelines stress consistency of various costs following mulit-view constraints. It is however, understandable, that NeRF is initially designed for view generation not for geometry.

## 6. CONCLUSIONS

This study presents a thorough evaluation of NeRF with comparison to three traditional MVS pipelines using two aerial photogrammetry datasets. Typically, NeRF was developed mostly at dealing with close-range cases with small to medium format images, often focusing on small scenes. Due to its high demand for both RAM and GPU memory, it presents computational challenges to process large-foramt and typical aerial photogrammetric blocks. To enable standard NeRF approach for such large-format images (i.e. 50-80 mega pixels), we presented a memory-efficient strategy to facilitate NeRF on large-scale 3D scene reconstruction. This approach reduces the memory demands by partitioning the training images into sub-image sets and employing efficient sampling techniques within the smaller sub-regions during the training process, and can adapt any NeRF methods to large-format data.

With adapting our proposed strategy to a typical NeRF structure such as Mip-NeRF, called our proposed NeRF method, we compared it against traditional MVS pipelines, and performed thorough experimental analysis to evaluate its potential to serve the photogrammetric 3D reconstruction purpose. Generally, we oberve that NeRF can recover the scene with better completes, with specifically outperform traditional MVS methods on reconstructing small objects (objects with small image pixel footprints). However, its performance on flat regions/large scenes are still not on par with typical MVS methods. This fact is due to that the NeRF structure and rendering process is not designed for geometry, where MVS method is solely solely designed for derive accurate geometry based on ray triangulating. More specific, and technical analysis these pros and cons can be found in Section 5.

The findings of this study shed light on on future research and improvement on NeRF to serve for photogrammetric 3D reconstruction purpose, as well means to incorporate its advantages into 3D reconstruction workflow. Firstly, the unique advantage of NeRF in reconstructing small objects, can be particularly useful and should be researched further for incorporating into photogrammetric 3D reconstruction process. Second, the depth rendering equation, and the intensity-based loss in NeRF are sub-optimal for 3D reconstruction, future endeavors can be valuable to improve them to be in favor of 3D reconstruction. Lastly, stressing multi-view consistency in the NeRF structure may further improve the 3D geometric reconstruction.

## VI. ACKNOWLEDGMENTS

The work was partially supported by the Office of Naval Research [grant numbers N00014-20-1-2141 & N00014-23-1-2670] and the project AI@TN funded by the Autonomous Province of Trento (Italy). The authors would like to thank ISPRS and EuroSDR for providing the Dortmund dataset and Leica/Hexagon for the Bordeaux dataset used in Section 4.